\documentclass[conference]{IEEEtran}
\IEEEoverridecommandlockouts

\AtBeginDocument{%
  \providecommand\BibTeX{{%
    \normalfont B\kern-0.5em{\scshape i\kern-0.25em b}\kern-0.8em\TeX}}}

\usepackage{graphicx}

\usepackage{colortbl}
\usepackage{amsmath}
\usepackage{mathtools}
\usepackage{diagbox}
\usepackage{pgfplots, pgfplotstable}
\usepackage{multirow}
\pgfplotsset{compat=1.11,
    /pgfplots/ybar legend/.style={
    /pgfplots/legend image code/.code={%
       \draw[##1,/tikz/.cd,yshift=-0.25em]
        (0cm,0cm) rectangle (3pt,0.8em);},
   },
}
\usepgfplotslibrary{colormaps}
\newenvironment{customlegend}[1][]{%
        \begingroup
        \csname pgfplots@init@cleared@structures\endcsname
        \pgfplotsset{#1}%
    }{%
        \csname pgfplots@createlegend\endcsname
        \endgroup
    }%
    \def\addlegendimage{\csname pgfplots@addlegendimage\endcsname}

\pgfplotsset{ 
cycle list={%
{draw=black,mark=star,solid},
{draw=black, mark=square,solid}}}

\usepackage{cite}
\usepackage{amsmath,amssymb,amsfonts}
\usepackage{algorithmic}
\usepackage{graphicx}
\usepackage{textcomp}
\usepackage{xcolor}
\usepackage[a4paper, total={180mm,250mm}]{geometry}
\def\BibTeX{{\rm B\kern-.05em{\sc i\kern-.025em b}\kern-.08em
    T\kern-.1667em\lower.7ex\hbox{E}\kern-.125emX}}

\usepackage{authblk}

\usepackage{caption}
\usepackage{booktabs}
\usepackage{graphicx}
\usepackage{colortbl}
\usepackage{amsmath}
\usepackage{mathtools}
\usepackage{diagbox}
\usepackage{pgfplots, pgfplotstable}
\usepackage{multirow}
\pgfplotsset{compat=1.11,
    /pgfplots/ybar legend/.style={
    /pgfplots/legend image code/.code={%
       \draw[##1,/tikz/.cd,yshift=-0.25em]
        (0cm,0cm) rectangle (3pt,0.8em);},
   },
}
\usepgfplotslibrary{colormaps}

\pgfplotsset{ 
cycle list={%
{draw=black,mark=star,solid},
{draw=black, mark=square,solid}}}

\usepackage{fancyhdr}
\fancypagestyle{firstpage}
{
    \fancyhead[L]{\footnotesize © 2024 IEEE.  Personal use of this material is permitted.  Permission from IEEE must be obtained for all other uses, in any current or future media, including reprinting/republishing this material for advertising or promotional purposes, creating new collective works, for resale or redistribution to servers or lists, or reuse of any copyrighted component of this work in other works. This paper is accepted at the 30th IEEE International Symposium on On-Line Testing and Robust System Design (IOLTS), 2024.}
    \fancyhead[R]{}
}

\begin{document}

\title{\huge \textbf{Cost-Effective Fault Tolerance for CNNs Using Parameter Vulnerability Based Hardening and Pruning}}

\author[1]{Mohammad Hasan Ahmadilivani}
\author[2]{Seyedhamidreza Mousavi}
\author[1]{\\Jaan Raik}
\author[1,2]{Masoud Daneshtalab}
\author[1]{Maksim Jenihhin}
\affil[1]{Tallinn University of Technology, Tallinn, Estonia}
\affil[2]{Mälardalen University, Västerås, Sweden}
\affil[1]{\{mohammad.ahmadilivani, jaan.raik, maksim.jenihhin\}@taltech.ee}
\affil[2]{\{seyedhamidreza.mousavi, masoud.daneshtalab\}@mdu.se}


\maketitle

\thispagestyle{firstpage}
 
\begin{abstract}

Convolutional Neural Networks (CNNs) have become integral in safety-critical applications, thus raising concerns about their fault tolerance. Conventional hardware-dependent fault tolerance methods, such as Triple Modular Redundancy (TMR), are computationally expensive, imposing a remarkable overhead on CNNs. Whereas fault tolerance techniques can be applied either at the hardware level or at the model levels, the latter provides more flexibility without sacrificing generality. This paper introduces a model-level hardening approach for CNNs by integrating error correction directly into the neural networks. The approach is hardware-agnostic and does not require any changes to the underlying accelerator device.
Analyzing the vulnerability of parameters enables the duplication of selective filters/neurons so that their output channels are effectively corrected with an efficient and robust correction layer. The proposed method demonstrates fault resilience nearly equivalent to TMR-based correction but with significantly reduced overhead. Nevertheless, there exists an inherent overhead to the baseline CNNs. To tackle this issue, a cost-effective parameter vulnerability based pruning technique is proposed that outperforms the conventional pruning method, yielding smaller networks with a negligible accuracy loss. Remarkably, the hardened pruned CNNs perform up to 24\% faster than the hardened un-pruned ones.


\end{abstract}



\section{Introduction} \label{sec:intro}
Convolutional Neural Networks (CNNs) have found widespread application in various safety-critical domains, owing to their superior accuracy compared to human performance \cite{krizhevsky2012imagenet,pouyanfar2018survey}.  
Hardware devices, including general-purpose processors (e.g., CPUs and GPUs) and specialized accelerators (e.g., FPGAs and ASICs), are employed to efficiently execute CNN models  ~\cite{chen2016eyeriss,jouppi2017datacenter}.
In many cases, especially with the general purpose accelerators but also with off-the-shelf integrated circuits and hard/firm cores, it is not possible to alter the underlying hardware to improve the fault tolerance of the CNN operation. 

The hardware systems deploying CNNs rely on extensive memory resources to store the parameters, making them susceptible to various fault effects due to transistor miniaturization \cite{canal2020predictive}.
Consequently, a major concern in deploying CNNs on hardware devices is their resilience to faults in memory, particularly those affecting their parameters. Extensive studies have demonstrated that faults in CNN parameters lead to drastic accuracy drops at very low error rates \cite{mittal2020survey,ahmadilivani2023Enhancing,ibrahim2020soft,su2023testability}.

\begin{figure}[h]
    \includegraphics[width=0.45\textwidth]{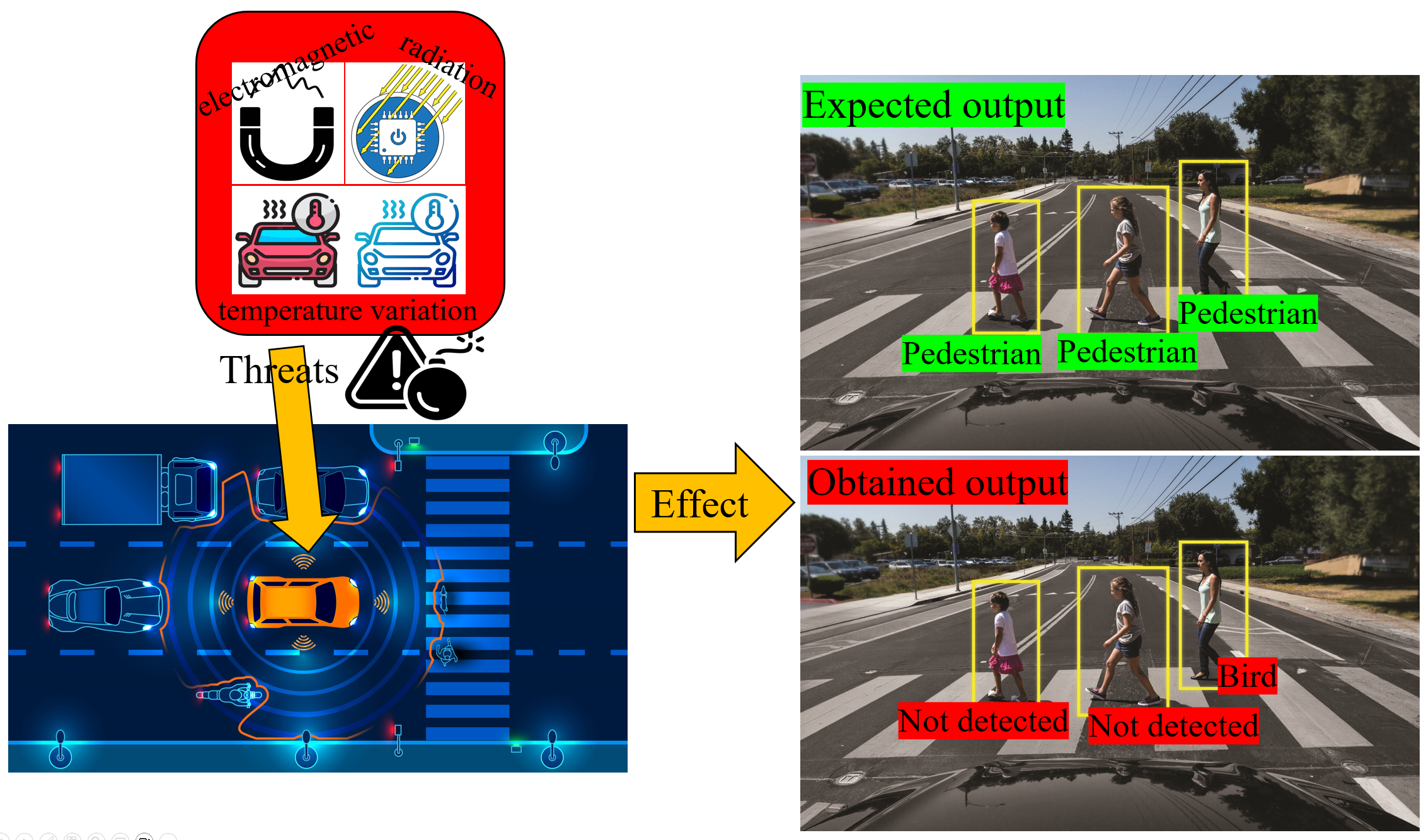}
    \centering
    \caption {Potential impact of faults on the output classification in the object detection task of an autonomous vehicle.}
    \label{fig:faults-effect}
\end{figure}

Fig. \ref{fig:faults-effect} illustrates an example of the effect of faults on the output classification in the object detection task of an autonomous vehicle. The faults can be a result of different causes, including temperature variation, terrestrial or cosmic radiation, circuit aging, or electromagnetic interference. A 
CNN running on a computing device in the vehicle classifies the input images and due to faults, some parameters are erroneous. As a result, the failure to recognize the pedestrians leads to a catastrophe. 
Therefore, it is crucial to enhance the fault tolerance of CNN models running on hardware devices to effectively employ them in safety-critical applications \cite{ahmadilivani2023systematic,ahmadilivani2023deepvigor,ahmadilivani2024special}.

To mitigate the impact of faults on the deployment of CNNs and at the same time avoid the high overhead in conventional fault-tolerant techniques such as Triple Modular Redundancy (TMR), researchers proposed selective hardening approaches \cite{schorn2018accurate,ruospo2021reliability,abdullah2020salvagednn,ahmadilivani2023Enhancing,libano2018selective}. Here, the objective is to protect the parameters or neurons that have a larger effect on the neural network's outputs against faults and errors. Therefore, the more vulnerable neurons are identified by resilience analysis and they are executed on hardened processing elements on the target hardware. 

Although these methods propose a model-level resilience analysis to identify the more vulnerable parameters/neurons, their protection techniques are restricted to FPGAs and ASICs that can be freely modified and redesigned. Whereas there exist numerous applications from high-performance to edge computing, where general-purpose computing devices such as CPUs and GPUs or hard and firm accelerator cores are deployed that do not support redesigning the hardware for fault-tolerance \cite{abich2022impact,zhao2020ft}. 

Moreover, \textit{fault-aware pruning} with retraining is another approach for improving fault tolerance of CNNs proposed by ~\cite{zhang2018analyzing,chitty2020model} in which the parameters that are mapped to corrupted processing elements of the target accelerator are pruned in the network. These methods are not only accelerator-specific but also should be applied to each individual chip with a different fault map separately. Moreover, they cannot be applied in the field but are designed to tolerate faults that have been already diagnosed in the laboratory. Thus, model-level fault tolerance approaches are preferred in terms of their flexibility.

\textit{Quantization} is shown to be highly effective for the resilience of CNNs \cite{ozen2021snr} since it restricts the numerical range within a CNN, thus eliminating the effect of large values produced due to faults and bitflips in a CNN. Nevertheless, apart from accuracy concerns, deploying quantized CNNs requires dedicated hardware accelerators for handling associated operations. Otherwise, they carry out the floating-point arithmetic of general-purpose computing devices \cite{gholami2022survey} which leads to the reliability issues of floating-point data types, that is contradictory to the purpose of hardening by quantization. Nonetheless, the model-level fault tolerance methods are mostly orthogonal to quantization and they can be employed on top of each other to improve the resilience of DNNs.

\textit{Fault-aware training} \cite{cavagnero2022fault,zahid2020fat} effectively improves the resilience of DNNs. However, it retrains the entire CNN with numerous fault injection scenarios that is not only excessively complex but also requires the possibility of having access to parameters. \textit{Error Correction Codes (ECC)} and \textit{Algorithm-based Fault Tolerance (ABFT)} utilize data encoding/decoding processes for real-time fault detection and correction ~\cite{zhao2020ft,lee2022value}. However, the practicality of these techniques in fault correction is questionable due to the overhead they introduce to memory and computations, posing a considerable challenge for CNNs that already have substantial memory and computational requirements.

\textit{Activation restriction} methods \cite{chen2021low,hoang2020ft,ghavami2022fitact} bound the activation values between layers through activation functions (i.e., ReLU) to mitigate error propagation to the outputs of CNNs. They clip the activations to 0 when their values exceed pre-identified ranges. These methods are effective in enhancing the resilience of CNNs, however, they do not provide error correction, and CNNs fail to work at high error rates due to the replacement of numerous feature maps with 0. \cite{ali2020erdnn} proposes a correction layer that executes each convolutional layer three times for fault detection and correction which however lays a prohibitive performance overhead to CNNs.

To overcome the previously mentioned issues, this paper introduces a novel model-level hardening solution to modify the architecture of CNNs to allow fault correction at inference inherently. An efficient error correction mechanism is designed enabled by selectively duplicated channels (in both convolutional and fully connected layers) within the structure of CNNs. In the proposed method, the parameter vulnerability of CNNs is analyzed and the more vulnerable ones are duplicated. Thereafter, a correction layer detects and corrects the erroneous output activations based on the two duplicated values.

The proposed hardening mechanism effectively reduces the overhead with respect to the TMR-based hardening solution, possessing the same fault tolerance capabilities. However, it still incurs some overhead to the memory and performance of the hardened CNN.
To further reduce this overhead, for the first time, a strategy is proposed for channel pruning based on the vulnerability of parameters to effectively shrink the size of CNNs with a negligible accuracy loss. 
In particular, we estimate the vulnerability of weight channels in CNNs, eliminate the least vulnerable ones to decrease the network's size, and then apply the hardening mechanism. The presented vulnerability-aware pruning provides the opportunity to eliminate any overhead caused by the protection mechanism on the designed hardened CNNs.


The contributions of this paper are as follows:
\begin{itemize}
    \item Proposing a model-level hardening method for CNNs to enhance their fault tolerance during inference. The approach involves duplicating the parameters in channels more vulnerable to faults and incorporating a highly effective Error Detection and Correction (EDAC) Layer to correct erroneous feature maps. 
    \item Proposing a channel pruning technique based on the parameter vulnerability that enables achieving a substantial reduction in the overhead incurred by the hardening mechanism.
    \item Results indicate that the proposed method allows hardened CNNs to perform reliably at error rates several orders of magnitude higher than those tolerated by the baseline CNN, achieved with merely 15\% selective parameter duplication. Moreover, leveraging pruning allows hardened pruned CNNs to be more resilient than the un-pruned ones, with up to 24\% higher performance in terms of execution time.
\end{itemize}

In the rest of the paper, Section \ref{sec:method} presents the proposed CNN model hardening through duplicated vulnerable channels and EDAC layer. Section \ref{sec:resilience-study} indicates the results achieved by the proposed method. In Section \ref{sec:pruning} the proposed parameter vulnerability based pruning is presented, and the overhead and resilience analyses are performed, and Section \ref{sec:conclusion} concludes the paper.

\section{CNN Model Hardening} \label{sec:method} 

In this section, the proposed hardening method to enhance the fault tolerance of CNN models is presented. This involves CNN architecture modification empowering them to inherently detect and correct faults. The method takes a pre-trained CNN and generates a hardened version that is executable by the target device. 


\subsection{Vulnerability Estimation} \label{subsec:vulnerability}

Vulnerability estimation of CNN's parameters reflects how they affect classification outputs in the presence of faults. Fault injection based approaches are very complex and time-consuming for addressing this task, whereas analytical approaches can estimate vulnerability fast and reasonably accurately \cite{ahmadilivani2023systematic}. This work adopts a vulnerability estimation approach introduced by \cite{mahmoud2020hardnn} and adapts it to the parameters of a channel in a CNN. This approach is accurate with fault injection results in \cite{mahmoud2020hardnn}. Eq. \eqref{eq:vulnerability-estimation} describes the vulnerability estimation for each channel:

\begin{equation}
    Vulnerability_{channel} = \sum_{i = 1, i \neq t}^{C} \frac{\sum_{w \in channel}|\frac{\partial(Z_i - Z_t)}{\partial w}|^2}{|Z_i - Z_t|^2}
    \label{eq:vulnerability-estimation}
\end{equation}

In Eq. \eqref{eq:vulnerability-estimation}, the \textit{vulnerability of a channel} with multiple weights \(w\) in a convolutional (CONV) layer of a CNN with \(C\) number of classes is estimated for a single input data. The output logits of the network corresponding to each output class is \(Z_i\) and the top class's logit is \(Z_t\). This equation represents the effect of each channel on the output logits as a vulnerability estimation and a higher value represents a higher vulnerability of the corresponding channel. A similar equation is applied to the weights corresponding to a neuron in Fully Connected (FC) layers.

\begin{figure*}[h!]
    \begin{center}
        \includegraphics[width=\textwidth]{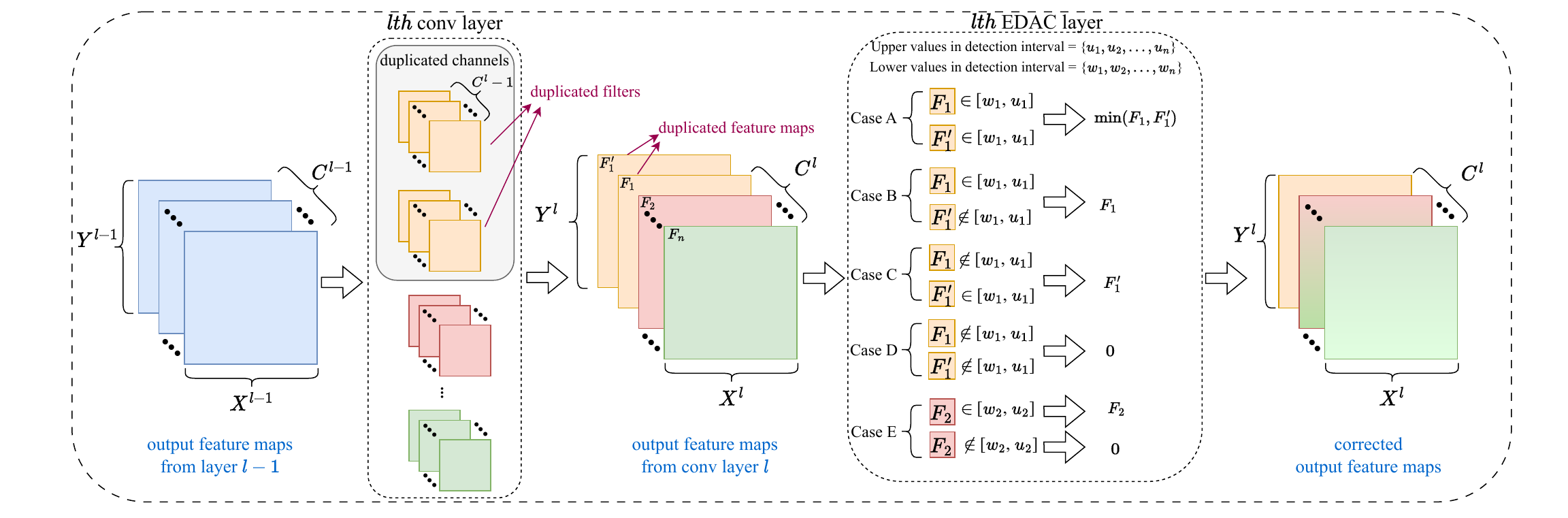}
        \caption{Channel duplication and EDAC layer}
        \label{fig:EDAC-method}
    \end{center}	
\end{figure*}

\subsection{CNN Model Hardening Method} \label{subsec:modification}

Subsequent to obtaining the vulnerability of channels of a pre-trained CNN, the CNN model is hardened by performing two steps:

\begin{itemize}
    \item Duplication of the more vulnerable channels, 
    \item Insertion of the Error Detection and Correction (EDAC) layer after each CONV/FC.
\end{itemize}

\subsubsection{\textbf{Channel Duplication}}

Fig. \ref{fig:EDAC-method} illustrates how the duplication of parameter channels functions. A channel contains multiple weights for obtaining an output feature map (fmap) \(F_k\) resulting from the summation of weighted inputs.  In the \(l\)th CONV layer with \(C^l\) output channels, a channel is a 3-dimensional array of weights \(X^l, Y^l, C^l\). (In an FC layer, an output channel is a 1-dimensional weight array corresponding to a neuron). Duplicating a channel of parameters generates duplicated values in \(F_k\) which provides an opportunity to detect and correct errors produced by faults in parameters. In this method, a ratio of more vulnerable channels with respect to Eq. \eqref{eq:vulnerability-estimation} are selected for duplication.




\subsubsection{\textbf{Error Detection And Correction (EDAC) Layer}}

After duplicating the vulnerable parameter channels, an EDAC layer is inserted into the CNN after each CONV and FC layer. The EDAC layer is meant to detect and correct errors in its incoming \(F_k\) from CONV/FC layers within the networks. One of the major challenges with 32-bit floating point data representation in general-purpose devices such as CPU and GPU is that faults may lead to overflows in CNNs producing Not-a-Number (NaN) values and corrupting the outputs. To address this issue, one of the primary operations in the EDAC layer is to replace any produced NaN value with 0 in the feature maps \(F_k\). 

Fig. \ref{fig:EDAC-method} illustrates how the EDAC layer operates. The EDAC layer exploits a detection interval containing the minimum and maximum values in the channels of \(F_k\) that are the lower values \(\{w_1, w_2, ..., w_n\}\) and the upper values \(\{u_1, u_2, ..., u_n\}\), respectively. Detection intervals are obtained by profiling the CNN on the training dataset. It is assumed that the data distribution of training is representative enough to provide generic and valid detection intervals for the unseen data during the inference \cite{Goodfellow2016}.

EDAC layer is aware of the duplicated and non-duplicated channels. In the duplicated channels, an error is detected and corrected in two cases: 
\begin{itemize}
    \item Both duplicated values in the corresponding channels are in the detection interval but are not equal. In this case, the minimum value between them is selected as the correct output \(F_k\) (case A in Fig. \ref{fig:EDAC-method}). The reason behind this correction is that CNNs are more resilient to small numbers \cite{hoang2020ft}.
  
    \item A value in a channel exceeds the detection interval, thus, the duplicated value that is in the detection interval is the correct value for the output \(F_k\) (case B and C in Fig. \ref{fig:EDAC-method}). If both duplicated values are not in the detection interval, the output \(F_k\) is set to 0 (case D in Fig. \ref{fig:EDAC-method}). 
\end{itemize}

In the non-duplicated channels, faults are detected and corrected based on the detection intervals. If any value in the channel exceeds the corresponding detection interval output \(F_k\) sets to 0 (case E in Fig. \ref{fig:EDAC-method}). The rationale behind zeroing is that it eliminates the propagation of erroneous values within a DNN. 
Note, that the detection and correction are repeated for each element of the two-dimensional array of the feature map \(F_k\).

To prevent faults from any immediate misclassification at the last layer, all output channels of the last layer in CNNs (i.e., neurons in the last FC layer) are duplicated and protected by an EDAC layer. It is worth mentioning that EDAC is implemented in a highly parallel way in Pytorch so that it can operate detection and correction on all duplicated and non-duplicated channels in parallel. Moreover, the hardened CNNs have the same accuracy as the baseline ones.

\section{Resilience and Overhead Results} \label{sec:resilience-study}

In this section, the results of fault injection experiments into the parameters of hardened CNNs are presented. 

\subsection{Fault Model} \label{subsec:fault-model}

The parameters of a pre-trained CNN could be faulty at inference time due to several reasons, including soft errors, temperature or voltage variation, process variation, aging, etc. To examine the resilience of CNNs, we model faults in the parameters by flipping their bits considering different Bit Error Rates (BERs). To this end, any layer in the CNN's parameters, including convolutional, Fully Connected (FC), batch normalization and EDAC layers is subject to a fault injection campaign. We have developed the fault injection on top of Pytorch, and the data representation is IEEE-754 32-bit floating point. The number of bitflips in a layer is equal to \(BER \times \#parameters \times 32\) in that layer. The fault injection simulations are performed on an NVIDIA 3090 GPU and any fault injection experiment is repeated 1000 times and the average accuracy drop is reported as the resilience metric. The experimented BERs are \(10^{-8}\), \(5 \times 10^{-8}\), \(10^{-7}\), \(5 \times 10^{-7}\), \(10^{-6}\), \(5 \times 10^{-6}\), \(10^{-5}\), \(5 \times 10^{-5}\), and \(10^{-4}\).

\subsection{Baseline CNNs} \label{subsec:cnns}

The experiments in this work are performed on three deep CNNs: AlexNet and VGG-11 trained on Cifar-10 and VGG-16 trained on Cifar-100. Their baseline accuracy as well as the number of parameters and MAC operations are reported in Table 1. The performance in terms of execution time of the CNNs over their test set is examined on an NVIDIA 3090 GPU coupled with an AMD Threadripper 3960X 24-core processor.

Note, that the accuracy of unprotected CNNs decreases drastically even at relatively low BERs. The unprotected AlexNet drops \(26\%\) at \(BER = 5\times 10^{-7}\) and the accuracy of unprotected VGG-11 and VGG16 drops \(24.07\%\) and  \(31.17\%\) at \(BER = 5\times 10^{-8}\), respectively. 

\begin{table}[h!]
\small
\centering
\caption{The baseline CNNs leveraged in this paper.}
\resizebox{\columnwidth}{!}{%
\begin{tabular}{cccccc}\toprule\toprule
CNN     & Dataset   & \begin{tabular}[c]{@{}c@{}} Base \\ accuracy \end{tabular}  & \#parameters & \#MACs & \begin{tabular}[c]{@{}c@{}}Performance \\ (sec)\end{tabular}    \\ \toprule\toprule
AlexNet & Cifar-10  & 73.15\%       & 21,623,562   & 42,316,288     & 0.591 \\ \toprule
VGG-11  & Cifar-10  & 92.85\%       & 9,228,362    & 153,293,824    & 0.655 \\ \toprule
VGG-16  & Cifar-100 & 73.20\%       & 34,015,396   & 332,756,992    & 0.782 \\ \bottomrule
\end{tabular}
}
\label{tab:cnns}
\end{table}

\subsection{Hardening by Channel Duplication vs. Triplication}

First, we demonstrate how EDAC performs if the detection intervals are not exploited for non-duplicated channels and compare it with a triplication-based correction performed by a voter. The voter takes three replicated fmaps in the corresponding channel and outputs the most repeated value. In the case where all three fmaps are different (if at least two replicated filters are faulty), the voter outputs the minimum value. 



Fig. \ref{fig:dmr-tmr-comparison} presents the results for accuracy drop and memory overhead of \textit{duplication + EDAC} vs. \textit{triplication + voter} for AlexNet at BER=\(10^{-4}\) over different channel hardening ratios. A similar trend is observed for VGG-11 and VGG-16. The highlights that can be observed from the Figure are:

\begin{itemize}
    \item \textit{Duplication + EDAC} achieves a similar resilience to that of \textit{triplication + voter} in terms of accuracy drop, with twice less memory overhead. 
    \item The memory overhead is proportional to the channel duplication and triplication ratio. The memory overhead of the EDAC layer is negligible compared to the total memory and computational requirements of CNNs.
    \item A high resilience is achieved only at full channel hardening. At lower hardening ratios, although the more vulnerable channels are protected, the unprotected channels incur a high accuracy drop in CNNs due to the high BER.
\end{itemize}

\begin{table}[h!]
\captionsetup{justification=centering}
\centering
\begin{tabular}{cc}
\multicolumn{2}{c}{
\begin{tikzpicture}
    \begin{customlegend}[legend columns=2,legend style={text opacity = 1,row sep=0pt, font=\fontsize{6}{4}\selectfont, column sep=1.5ex},
        legend entries={{Dup. + EDAC},
                        {Trip. + voter}}]
        \addlegendimage{mark=triangle*, mark size=2pt, blue, thick ,draw=blue}
        \addlegendimage{mark=square, mark size=2pt, red, thick,draw=red}
    \end{customlegend}
\end{tikzpicture}}

\\

\begin{tikzpicture}
\pgfplotsset{
  log x ticks with fixed point/.style={
      xticklabel={
        \pgfkeys{/pgf/fpu=true}
        \pgfmathparse{exp(\tick)}%
        \pgfmathprintnumber[fixed relative, precision=3]{\pgfmathresult}
        \pgfkeys{/pgf/fpu=false}
      }
  },
  log y ticks with fixed point/.style={
      yticklabel={
        \pgfkeys{/pgf/fpu=true}
        \pgfmathparse{exp(\tick)}%
        \pgfmathprintnumber[fixed relative, precision=3]{\pgfmathresult}
        \pgfkeys{/pgf/fpu=false}
      }
  }
}
  \begin{axis}[xmode=log,
        width=0.5\columnwidth,
        height=0.4\columnwidth,
        font=\footnotesize,
        scaled x ticks = false,
        scaled y ticks = false,
        xtick={90, 95, 100},
        ytick={10, 30, 50, 70},
        xticklabels = {\strut $90$, \strut $95$, \strut $100$},
        yticklabels = {\strut $10$, \strut $30$,\strut $50$,\strut $70$},
        ymin=0, ymax=70,
        xmin=90, xmax=100,
        grid=major, 
        grid style={dashed,gray}, 
        ylabel near ticks,
        xlabel near ticks,
        xlabel= Channel hardening ratio (\%), 
        ylabel= Accuracy drop (\%),
        legend columns = 2,
        legend style={at={(1,1)},anchor=north},
        legend style={draw=black, at={(0.5,-0.5)}, text opacity = 1,row sep=0pt, font=\fontsize{6}{4}\selectfont},
         x tick label style={rotate=0,anchor=north},
        ]
        \addplot [blue, thick,mark=triangle*,mark size=2pt] table [x=protection, y=BD5, col sep=comma] {charts/alexnet-dmr-resilience.csv};
        \addplot [red, thick ,mark=square , mark size=2pt] table [x=protection, y=BT5, col sep=comma] {charts/alexnet-dmr-resilience.csv};
      \end{axis}
    \end{tikzpicture}
   
   & 

\begin{tikzpicture}   
\pgfplotsset{
  log x ticks with fixed point/.style={
      xticklabel={
        \pgfkeys{/pgf/fpu=true}
        \pgfmathparse{exp(\tick)}%
        \pgfmathprintnumber[fixed relative, precision=3]{\pgfmathresult}
        \pgfkeys{/pgf/fpu=false}
      }
  },
  log y ticks with fixed point/.style={
      yticklabel={
        \pgfkeys{/pgf/fpu=true}
        \pgfmathparse{exp(\tick)}%
        \pgfmathprintnumber[fixed relative, precision=3]{\pgfmathresult}
        \pgfkeys{/pgf/fpu=false}
      }
  }
}
  \begin{axis}[xmode=log,
        width=0.5\columnwidth,
        height=0.4\columnwidth,
        font=\footnotesize,
        scaled x ticks = false,
        scaled y ticks = false,
        xtick={90, 95, 100},
        ytick={0, 50, 100, 150, 200},
        xticklabels = {\strut $90$, \strut $95$, \strut $100$},
        yticklabels = {\strut $0$, \strut $50$, \strut $100$,\strut $150$,\strut $200$},
        ymin=0, ymax=210,
        xmin=90, xmax=100,
        grid=major, 
        grid style={dashed,gray}, 
        ylabel near ticks,
        xlabel near ticks,
        xlabel= Channel hardening ratio (\%), 
        ylabel= Memory overhead (\%),
        legend columns = 2,
        legend style={at={(1,1)},anchor=north},
        legend style={draw=black, at={(0.5,-0.5)}, text opacity = 1,row sep=0pt, font=\fontsize{6}{4}\selectfont},
         x tick label style={rotate=0,anchor=north},
        ]
        \addplot [blue, thick ,mark=triangle* , mark size=2pt] table [x=protection, y=DMR parameters, col sep=comma] {charts/alexnet-dmr-overhead_1.csv};
        \addplot [red, thick ,mark=square , mark size=2pt] table [x=protection, y=TMR parameters, col sep=comma] {charts/alexnet-dmr-overhead_1.csv};
      \end{axis}
    \end{tikzpicture}

\\

\;\;\;\;\;\;\;\;\;  (a) Resilience &   \;\;\;\;\;\;\;\;   (b)  Memory Overhead 

\end{tabular}

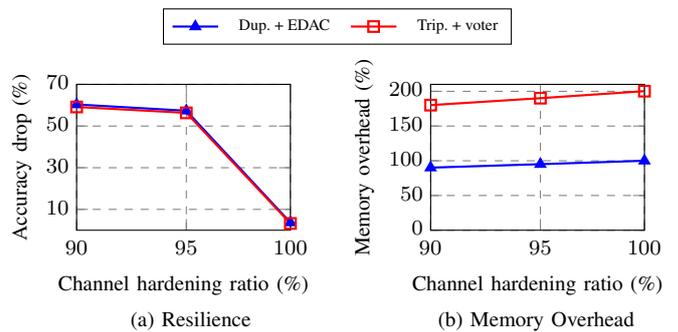
\captionof{figure}{Resilience (a) and memory overhead (b) for AlexNet hardened by \textit{duplication + EDAC} vs. \textit{triplication + voter} at BER=\(10^{-4}\), without applying detection intervals to Non-duplicated channels.
}
\label{fig:dmr-tmr-comparison}
\end{table}

As observed, we need to apply a full channel duplication + EDAC to protect CNNs which leads to a significant overhead in the hardened CNNs compared to the unprotected ones. The hardened CNNs have double memory and computational requirements (\(100\%\) overhead) and the execution time increases up to \(1.83\) times. To tackle this issue, we exploit the detection intervals in the non-duplicated channels to protect less vulnerable channels which leads to lower hardening ratios. It is presented in the next subsection.

\subsection{Hardening by Selective Channels Duplication and EDAC Layer}

In this subsection, we present the results for the selective channel duplication with EDAC layers as described in Fig. \ref{fig:EDAC-method}. In a pre-trained DNN, a ratio of the more vulnerable channels are duplicated and both duplicated and non-duplicated channels exploit detection intervals to be hardened at EDAC layer. Since the hardening method is at the model level, the performance in terms of execution time is influenced. Thus, we present the performance overhead on NVIDIA 3090 GPU in the paper. 

\begin{table*}[h!]
\captionsetup{justification=centering}
\centering
\resizebox{\textwidth}{!}{
\begin{tabular}{c|c|c}
\multicolumn{3}{c}{
\begin{tikzpicture}
    \begin{customlegend}[legend columns=3,legend style={text opacity = 1,row sep=0pt, font=\fontsize{6}{4}\selectfont, column sep=1.5ex},
        legend entries={{Accuracy drop},
                        {Performance Overhead}}]
        \addlegendimage{mark=triangle*, mark size=2pt, blue, thick ,draw=blue}
        \addlegendimage{mark=square, mark size=2pt, red, thick,draw=red}
    \end{customlegend}
\end{tikzpicture}}
\\
\begin{tikzpicture}
    \pgfplotsset{
      log x ticks with fixed point/.style={
          xticklabel={
            \pgfkeys{/pgf/fpu=true}
            \pgfmathparse{exp(\tick)}%
            \pgfmathprintnumber[fixed relative, precision=3]{\pgfmathresult}
            \pgfkeys{/pgf/fpu=false}
          }
      },
      log y ticks with fixed point/.style={
          yticklabel={
            \pgfkeys{/pgf/fpu=true}
            \pgfmathparse{exp(\tick)}%
            \pgfmathprintnumber[fixed relative, precision=3]{\pgfmathresult}
            \pgfkeys{/pgf/fpu=false}
          }
      }
    }
    \begin{axis}[
        width=0.3\textwidth,
        height=0.5\columnwidth,
        font=\footnotesize,
        scaled x ticks = false,
        scaled y ticks = false,
        xtick={0, 5, 10, 15},
        ytick={3, 3.15, 3.3, 3.45,3.6},
        xticklabels = {\strut $0$, \strut $5$, \strut $10$, \strut $15$},
        yticklabels = {\strut $3.00$, \strut $3.15$, \strut $3.30$,\strut $3.45$, \strut $3.60$},
        axis y line*=left,
        ymin=3, ymax=3.6,
        xmin=0, xmax=15,
        grid=major, 
        grid style={dashed,gray},
        ylabel near ticks,
        xlabel near ticks,
        xlabel=Channel hardening ratio (\%),
        ylabel=Accuracy drop (\%),
        ]
        \addplot [blue, thick,mark=triangle*,mark size=2pt] table [x=protection, y=accuracy, col sep=comma] {charts/alexnet-edac-resilience-performance_1.csv};  
    \end{axis}

    \begin{axis}[
        width=0.3\textwidth,
        height=0.5\columnwidth,
        font=\footnotesize,
        scaled y ticks = false,
        xtick={0, 5, 10, 15},
        ytick={4, 7.25, 10.5, 13.75,17},
        xticklabels = {\strut $0$, \strut $5$, \strut $10$, \strut $15$},
        yticklabels = {\strut $4.00$,\strut $7.25$,\strut $10.50$,\strut $13.75$ ,\strut $17.00$},
        axis y line*=left,
        ymin=4, ymax=17,
        xmin=0, xmax=15,
        grid=major, 
        grid style={dashed,gray},
        ylabel near ticks,
        xlabel near ticks,
        axis y line*=right,
        axis x line=none,
        ylabel=Performance overhead (\%)
        ]
        \addplot [red, thick ,mark=square , mark size=2pt] table [x=protection, y=performance, col sep=comma] {charts/alexnet-edac-resilience-performance_1.csv}; 
    \end{axis}
\end{tikzpicture}
   
   & 

\begin{tikzpicture}
    \pgfplotsset{
      log x ticks with fixed point/.style={
          xticklabel={
            \pgfkeys{/pgf/fpu=true}
            \pgfmathparse{exp(\tick)}%
            \pgfmathprintnumber[fixed relative, precision=3]{\pgfmathresult}
            \pgfkeys{/pgf/fpu=false}
          }
      },
      log y ticks with fixed point/.style={
          yticklabel={
            \pgfkeys{/pgf/fpu=true}
            \pgfmathparse{exp(\tick)}%
            \pgfmathprintnumber[fixed relative, precision=3]{\pgfmathresult}
            \pgfkeys{/pgf/fpu=false}
          }
      }
    }
    \begin{axis}[
        width=0.3\textwidth,
        height=0.5\columnwidth,
        font=\footnotesize,
        scaled x ticks = false,
        scaled y ticks = false,
        xtick={0, 5, 10, 15},
        ytick={2.8, 3.1,3.4, 3.7, 4},
        xticklabels = {\strut $0$, \strut $5$, \strut $10$, \strut $15$},
        yticklabels = {\strut $2.8$, \strut $3.1$,\strut $3.4$, \strut $3.7$, \strut $4.0$},
        axis y line*=left,
        ymin=2.8, ymax=4,
        xmin=0, xmax=15,
        grid=major, 
        grid style={dashed,gray},
        ylabel near ticks,
        xlabel near ticks,
        xlabel=Channel hardening ratio (\%),
        ylabel=Accuracy drop (\%),
        ]
        \addplot [blue, thick,mark=triangle*,mark size=2pt] table [x=protection, y=accuracy, col sep=comma] {charts/vgg11-edac-resilience-performance_1.csv};  
    \end{axis}

    \begin{axis}[
        width=0.3\textwidth,
        height=0.5\columnwidth,
        font=\footnotesize,
        scaled y ticks = false,
        xtick={0, 5, 10, 15},
        ytick={8.5, 11,13.5,16,18.5},
        xticklabels = {\strut $0$, \strut $5$, \strut $10$, \strut $15$},
        yticklabels = {\strut $8.5$, \strut $11.0$,\strut $13.5$,\strut $16.0$,\strut $18.5$},
        axis y line*=left,
        ymin=8.5, ymax=18.5,
        xmin=0, xmax=15,
        grid=major, 
        grid style={dashed,gray},
        ylabel near ticks,
        xlabel near ticks,
        axis y line*=right,
        axis x line=none,
        ylabel=Performance overhead (\%)
        ]
        \addplot [red, thick ,mark=square , mark size=2pt] table [x=protection, y=performance, col sep=comma] {charts/vgg11-edac-resilience-performance_1.csv}; 
    \end{axis}
\end{tikzpicture}

&

\begin{tikzpicture}
    \pgfplotsset{
      log x ticks with fixed point/.style={
          xticklabel={
            \pgfkeys{/pgf/fpu=true}
            \pgfmathparse{exp(\tick)}%
            \pgfmathprintnumber[fixed relative, precision=3]{\pgfmathresult}
            \pgfkeys{/pgf/fpu=false}
          }
      },
      log y ticks with fixed point/.style={
          yticklabel={
            \pgfkeys{/pgf/fpu=true}
            \pgfmathparse{exp(\tick)}%
            \pgfmathprintnumber[fixed relative, precision=3]{\pgfmathresult}
            \pgfkeys{/pgf/fpu=false}
          }
      }
    }
    \begin{axis}[
        width=0.3\textwidth,
        height=0.5\columnwidth,
        font=\footnotesize,
        scaled x ticks = false,
        scaled y ticks = false,
        xtick={0, 5, 10, 15},
        ytick={1.6, 1.7, 1.8, 1.9, 2,2.1},
        xticklabels = {\strut $0$, \strut $5$, \strut $10$, \strut $15$},
        yticklabels = {\strut $1.6$, \strut $1.7$,\strut $1.8$,\strut $1.9$, \strut $2.0$,\strut $2.1$},
        axis y line*=left,
        ymin=1.6, ymax=2.1,
        xmin=0, xmax=15,
        grid=major, 
        grid style={dashed,gray},
        ylabel near ticks,
        xlabel near ticks,
        xlabel=Channel hardening ratio (\%),
        ylabel=Accuracy drop (\%),
        ]
        \addplot [blue, thick,mark=triangle*,mark size=2pt] table [x=protection, y=accuracy, col sep=comma] {charts/vgg16-edac-resilience-performance.csv};  
    \end{axis}

    \begin{axis}[
        width=0.3\textwidth,
        height=0.5\columnwidth,
        font=\footnotesize,
        scaled y ticks = false,
        xtick={0, 5, 10, 15},
        ytick={11, 12.6, 14.2, 15.8,17.4,19},
        xticklabels = {\strut $0$, \strut $5$, \strut $10$, \strut $15$},
        yticklabels = {\strut $11.0$, \strut $12.6$,\strut $14.2$,\strut $15.8$,\strut $17.4$,\strut $19.0$},
        axis y line*=left,
        ymin=11, ymax=19,
        xmin=0, xmax=15,
        grid=major, 
        grid style={dashed,gray},
        ylabel near ticks,
        xlabel near ticks,
        axis y line*=right,
        axis x line=none,
        ylabel=Performance overhead (\%)
        ]
        \addplot [red, thick ,mark=square , mark size=2pt] table [x=protection, y=performance, col sep=comma] {charts/vgg16-edac-resilience-performance_1.csv}; 
    \end{axis}
\end{tikzpicture}

\\

\;\;\;\;\;\;\;  (a) AlexNet (BER=\(10^{-4}\)) &   \;\;\;\;\;\;   (b) VGG-11 (BER=\(5\times 10^{-5}\)) &   \;\;\;\;\;\;   (c) VGG-16 (BER=\(10^{-5}\))

\end{tabular}
}

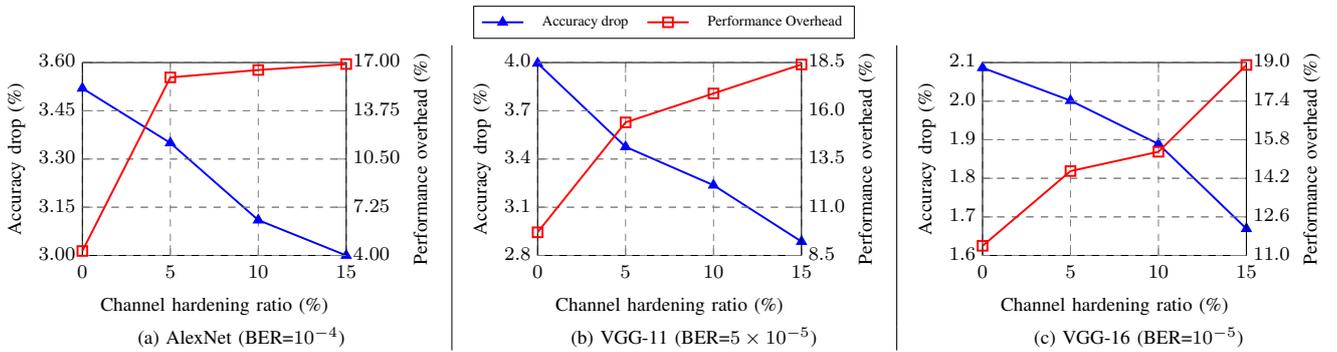
\captionof{figure}{Accuracy loss and performance overhead comparison for hardened AlexNet (a), hardened VGG-11 (b), and hardened VGG-16 (c), over different channel pruning ratios at the corresponding BERs where accuracy drop is lower than 5\%.}
\label{fig:edac-acc-perf-comparison}
\end{table*}



Fig. \ref{fig:edac-acc-perf-comparison} demonstrates the resilience and performance overhead for all the experimented CNNs at the highest BERs where the accuracy drop is not yet significant (lower than 5\%). 
As observed, exploiting detection intervals in unprotected channels has a remarkable effect on reducing the hardening ratio to achieve high resilience. 
It can be observed that by merely using detection intervals without channel duplication (\textit{hardening ratio = 0\%}), the accuracy drops at high BERs are lower than 4\% with up to 11.4\% performance overhead compared to the unprotected baseline CNNs. Nonetheless, increasing the channel hardening ratio improves the resilience without a significant performance overhead.

With a 15\% channel hardening ratio 
the accuracy drop improves 17\%, 38\%, and 24\% for AlexNet, VGG-11, and VGG-16 respectively, achieved by 6.7\% to 12\% longer execution time, compared to 0\% hardening ratio.

As noted, EDAC layers exploiting detection intervals for all channels can significantly reduce the overhead of the hardened CNNs compared to the full duplication. However, a tangible overhead is incurred to CNNs due to hardening. The overheads are caused by both channel duplication and EDAC layer operations. To tackle this issue, we deploy a pruning method to reduce the size of baseline CNNs by removing the least vulnerable channels and applying EDAC to the most vulnerable ones, so the total overhead can be further reduced. This method is presented in the next section.

\section{Overhead Reduction by Parameter Vulnerability Based Pruning} \label{sec:pruning}

As observed, although the introduced hardening technique exhibits a high resilience to CNNs, it lays a considerable overhead to them. To address this issue, we apply an effective structured channel pruning to CNNs to shrink their baseline size and open room for the hardening mechanism. 

\subsection{Vulnerability Based Pruning}

Structured pruning is a well-known method for CNN models to reduce their size leading to optimizing their performance and resource utilization. In this method, a metric for the significance of the effect of parameters on the output accuracy is considered and the least important weights are removed from the CNN with a negligible accuracy loss.

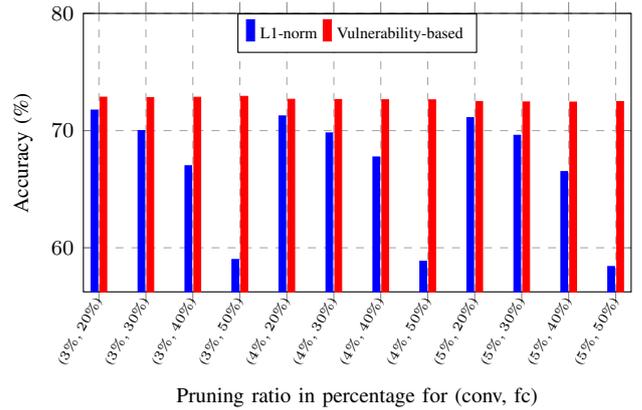
\begin{figure}[h!]
\captionsetup{justification=centering}
\centering
\begin{tikzpicture}
    \begin{axis}[
        font=\footnotesize,
        ybar=0.03cm,
        ymax=80,
        xticklabels={($3\%, 20\%$), ($3\%, 30\%$), ($3\%, 40\%$), ($3\%, 50\%$), ($4\%, 20\%$), ($4\%, 30\%$), ($4\%, 40\%$), ($4\%, 50\%$), ($5\%, 20\%$), ($5\%, 30\%$), ($5\%, 40\%$), ($5\%, 50\%$)},
        width=\columnwidth,
        height=0.6\columnwidth,
        ylabel near ticks,
        xlabel near ticks,
        grid=major,
        grid style={dashed,gray!70},
        ylabel= {{Accuracy (\%)}},
        xlabel= {{Pruning ratio in percentage for (conv, fc)}},
        tick label style={font=\footnotesize},  
        enlarge x limits=0.03,
        legend style={at={(0.5,1)},
            anchor=north,legend columns=-1, font=\fontsize{6}{4}\selectfont},
      x tick label style={rotate=60,anchor=north,font=\tiny, xshift=-2.8ex , yshift = 1.5ex},
        xtick=data,
        nodes near coords align={vertical},
        bar width=2.5pt
    ]
    \addplot [blue, fill] %
        coordinates {(1,71.75) (2,70) (3,67) (4,59) (5,71.25) (6,69.8) (7,67.75) (8,58.85) (9,71.1) (10,69.6) (11,66.5) (12,58.4)};
    \addplot [red, fill] %
        coordinates {(1,72.85) (2,72.81) (3,72.84) (4,72.92) (5,72.67) (6,72.65) (7,72.63) (8,72.62) (9,72.48) (10,72.44) (11,72.43) (12,72.48)};

    \legend{L1-norm, Vulnerability-based}
    \end{axis}
\end{tikzpicture}
 \caption{Comparison of L1-norm pruning and vulnerability-based pruning in AlexNet.}
\label{fig:pruning-cmp-alexnet}
\end{figure}

Conventionally, the significance of the weights effect is examined by L1-norm which is shown to be effective \cite{han2015deep}. In this work, we exploit Eq. \eqref{eq:vulnerability-estimation} as the importance metric for channel parameters and remove a ratio of the least vulnerable channels from CONV and FC layers in CNNs. To avoid losing too much accuracy, we perform lightweight training on the pruned CNNs with 10 epochs using SGD with a learning rate of 0.001 on the training dataset. Fig. \ref{fig:pruning-cmp-alexnet} shows that our vulnerability-aware pruning method is more effective than L1-norm pruning in terms of removing the channels of CNNs while the accuracy is still close to that of the baseline CNN. 

To obtain the highest possible pruning ratios for each CNN, we perform an extensive exploration over different pruning ratios of CONV and FC layers to minimize the number of parameters and MAC operations maintaining the test accuracy within 1\% of its unprotected baseline. 
Table \ref{tab:pruned-cnns} shows the selected pruning ratios for the experimented CNNs and their improved memory and computational requirements compared to the baseline ones. As it is observed, the pruned CNNs achieve from 1.18 to 6.19 times fewer parameters, 1.03 to 2.06 times fewer MAC operations, and 1\% to 11.1\% less execution time than the baseline ones.

\begin{table}[h!]
\small
\centering
\caption{Pruning ratio and normalized number of parameters and MAC operations and performance for each CNN.}
\resizebox{\columnwidth}{!}{

\begin{tabular}{ccccccc}
\hline
CNNs           & \begin{tabular}[c]{@{}c@{}}Conv. \\ prun. ratio\end{tabular} & \begin{tabular}[c]{@{}c@{}}FC prun. \\ ratio\end{tabular} & \begin{tabular}[c]{@{}c@{}}Pruned CNN \\ Accuracy\end{tabular} & \begin{tabular}[c]{@{}c@{}}Norm. \\ \#params to \\  baseline\end{tabular} & \begin{tabular}[c]{@{}c@{}}Norm. \\ \#MACs to \\  baseline\end{tabular} & \begin{tabular}[c]{@{}c@{}}Norm. \\ perf. to \\ baseline\end{tabular} \\ \toprule
 AlexNet  & 5\%  & 80\%  & 72.38\%  & 0.1615     & 0.4851  & 0.888
\\ \toprule
VGG-11 & 4\%   & 35\%    & 91.96\%     & 0.847    & 0.9059   &  0.987
\\ \toprule
VGG-16 & 1\%   & 15\%    & 72.4\%    & 0.826   & 0.9665  &   0.998
\\ \bottomrule
\end{tabular}
}
\label{tab:pruned-cnns}
\end{table}

\begin{table*}[t!]
\captionsetup{justification=centering}
\centering
\resizebox{\textwidth}{!}{
\begin{tabular}{c|c|c}
\multicolumn{3}{c}{
\begin{tikzpicture}
    \begin{customlegend}[legend columns=2,legend style={text opacity = 1,row sep=0pt, font=\fontsize{6}{4}\selectfont, column sep=1.5ex},
        legend entries={{Hardened baseline},
                        {Hardened pruned}}]
        \addlegendimage{mark=triangle*, mark size=2pt, blue, thick ,draw=blue}
        \addlegendimage{mark=square, mark size=2pt, red, thick,draw=red}
    \end{customlegend}
\end{tikzpicture}}
\\
\begin{tikzpicture}
\pgfplotsset{
  log x ticks with fixed point/.style={
      xticklabel={
        \pgfkeys{/pgf/fpu=true}
        \pgfmathparse{exp(\tick)}%
        \pgfmathprintnumber[fixed relative, precision=3]{\pgfmathresult}
        \pgfkeys{/pgf/fpu=false}
      }
  },
  log y ticks with fixed point/.style={
      yticklabel={
        \pgfkeys{/pgf/fpu=true}
        \pgfmathparse{exp(\tick)}%
        \pgfmathprintnumber[fixed relative, precision=3]{\pgfmathresult}
        \pgfkeys{/pgf/fpu=false}
      }
  }
}
  \begin{axis}[xmode=log,
        width=0.33\textwidth,
        height=0.5\columnwidth,
        font=\footnotesize,
        xtick={0.000005, 0.00001, 0.00005, 0.0001},
        ytick={0, 1, 2, 3, 4},
        xticklabels = {\strut $5 \times 10^{-6}$, \strut $10^{-5}$, \strut $5 \times 10^{-5}$, \strut $10^{-4}$},
        yticklabels = {\strut $0$,\strut $1$, \strut $2$,\strut $3$},
        ymin=0, ymax=3.3,
        xmin=0.000005, xmax=0.0001,
        grid=major, 
        grid style={dashed,gray}, 
        xlabel= BER, 
        ylabel= Accuracy drop (\%),
        legend columns = 2,
        legend style={at={(1,1)},anchor=north},
        legend style={draw=black, at={(0.5,-0.5)}, text opacity = 1,row sep=0pt, font=\fontsize{6}{4}\selectfont},
         x tick label style={rotate=20,anchor=north},
        ]
        \addplot [blue, thick,mark=triangle*,mark size=2pt] table [x=BER, y=baseline-protected, col sep=comma] {charts/pruned-alexnet-resilience.csv};
        \addplot [red, thick ,mark=square , mark size=2pt] table [x=BER, y=pruned-protected, col sep=comma] {charts/pruned-alexnet-resilience.csv};
      \end{axis}
    \end{tikzpicture}
   
   &

\begin{tikzpicture}
\pgfplotsset{
  log x ticks with fixed point/.style={
      xticklabel={
        \pgfkeys{/pgf/fpu=true}
        \pgfmathparse{exp(\tick)}%
        \pgfmathprintnumber[fixed relative, precision=3]{\pgfmathresult}
        \pgfkeys{/pgf/fpu=false}
      }
  },
  log y ticks with fixed point/.style={
      yticklabel={
        \pgfkeys{/pgf/fpu=true}
        \pgfmathparse{exp(\tick)}%
        \pgfmathprintnumber[fixed relative, precision=3]{\pgfmathresult}
        \pgfkeys{/pgf/fpu=false}
      }
  }
}
  \begin{axis}[xmode=log,
        width=0.33\textwidth,
        height=0.5\columnwidth,
        font=\footnotesize,
        xtick={0.000001, 0.000005, 0.00001, 0.00005},
        ytick={0, 1, 2, 3, 4},
        xticklabels = {\strut $10^{-6}$, \strut $5 \times 10^{-6}$, \strut $10^{-5}$, \strut $5 \times 10^{-5}$},
        yticklabels = {\strut $0$, \strut $1$, \strut $2$,\strut $3$},
        ymin=0, ymax=3.3,
        xmin=0.000001, xmax=0.00005,
        grid=major, 
        grid style={dashed,gray}, 
        ylabel near ticks,
        xlabel near ticks,
        xlabel= BER, 
        ylabel= Accuracy drop (\%),
        legend columns = 2,
        legend style={at={(1,1)},anchor=north},
        legend style={draw=black, at={(0.5,-0.5)}, text opacity = 1,row sep=0pt, font=\fontsize{6}{4}\selectfont},
         x tick label style={rotate=20,anchor=north},
        ]
        \addplot [blue, thick,mark=triangle*,mark size=2pt] table [x=BER, y=baseline-protected, col sep=comma] {charts/pruned-vgg11-resilience.csv};
        \addplot [red, thick ,mark=square , mark size=2pt] table [x=BER, y=pruned-protected, col sep=comma] {charts/pruned-vgg11-resilience.csv};
      \end{axis}
    \end{tikzpicture}

&

\begin{tikzpicture}
\pgfplotsset{
  log x ticks with fixed point/.style={
      xticklabel={
        \pgfkeys{/pgf/fpu=true}
        \pgfmathparse{exp(\tick)}%
        \pgfmathprintnumber[fixed relative, precision=3]{\pgfmathresult}
        \pgfkeys{/pgf/fpu=false}
      }
  },
  log y ticks with fixed point/.style={
      yticklabel={
        \pgfkeys{/pgf/fpu=true}
        \pgfmathparse{exp(\tick)}%
        \pgfmathprintnumber[fixed relative, precision=3]{\pgfmathresult}
        \pgfkeys{/pgf/fpu=false}
      }
  }
}
  \begin{axis}[xmode=log,
        width=0.33\textwidth,
        height=0.5\columnwidth,
        font=\footnotesize,
        scaled x ticks = false,
        scaled y ticks = false,
        xtick={0.000001, 0.000005, 0.00001, 0.00005},
        ytick={0, 3, 6, 9, 12},
        xticklabels = {\strut $10^{-6}$, \strut $5 \times 10^{-6}$, \strut $10^{-5}$, \strut $5 \times 10^{-5}$},
        yticklabels = {\strut $0$, \strut $3$, \strut $6$,\strut $9$},
        ymin=0, ymax=10,
        xmin=0.000001, xmax=0.00005,
        grid=major, 
        grid style={dashed,gray}, 
        ylabel near ticks,
        xlabel near ticks,
        xlabel= BER, 
        ylabel= Accuracy drop (\%),
        legend columns = 2,
        legend style={at={(1,1)},anchor=north},
        legend style={draw=black, at={(0.5,-0.5)}, text opacity = 1,row sep=0pt, font=\fontsize{6}{4}\selectfont},
         x tick label style={rotate=20,anchor=north},
        ]
        \addplot [blue, thick,mark=triangle*,mark size=2pt] table [x=BER, y=baseline-protected, col sep=comma] {charts/pruned-vgg16-resilience.csv};
        \addplot [red, thick ,mark=square , mark size=2pt] table [x=BER, y=pruned-protected, col sep=comma] {charts/pruned-vgg16-resilience.csv};
      \end{axis}
    \end{tikzpicture}

\\

\;\;\;\;\;\;\;\;\;  (a) AlexNet &   \;\;\;\;\;\;\;\;   (b) VGG-11 &   \;\;\;\;\;\;\;\;   (c) VGG-16 

\end{tabular}
}

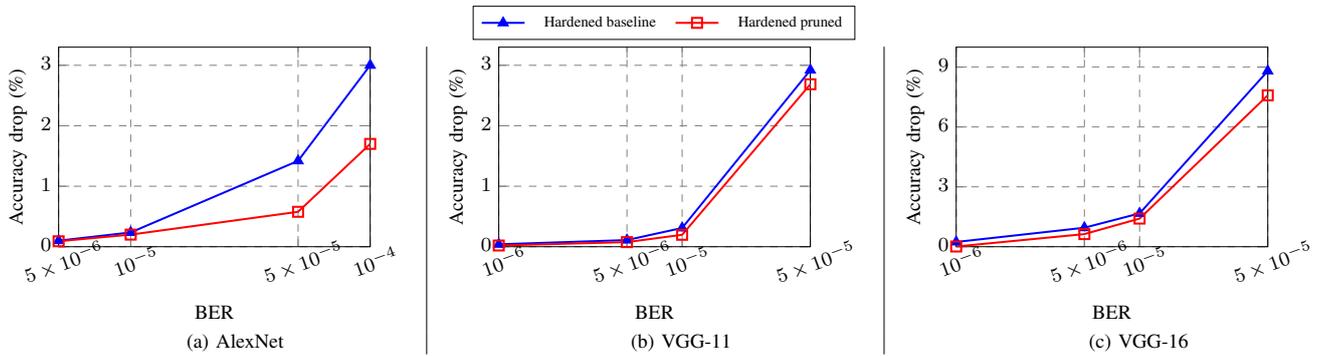
\captionof{figure}{Resilience comparison in terms of accuracy drop of hardened baseline and hardened pruned CNNs at different BERs with 15\% channel hardening ratio.}
\label{fig:pruned-resilience}
\end{table*}

\subsection{Resilience and Overhead Study of the Hardened Pruned CNNs}

By shrinking the baseline CNNs using pruning, we have the opportunity to minimize the overhead of hardened CNNs compared to the baseline ones. Now, the pruned pre-trained CNNs are hardened by the method introduced in Section \ref{sec:method}. Their channel vulnerability is obtained, the more vulnerable channels are duplicated, and EDAC layers are implanted into the model with the corresponding detection intervals. 

Fig. \ref{fig:pruned-resilience} illustrates how resilience is improved in the hardened pruned CNNs against hardened baseline ones over different BERs, with 15\% channel hardening ratio. It is observed that the proposed pruning not only reduces the overhead of hardened CNNs but also improves their resilience.

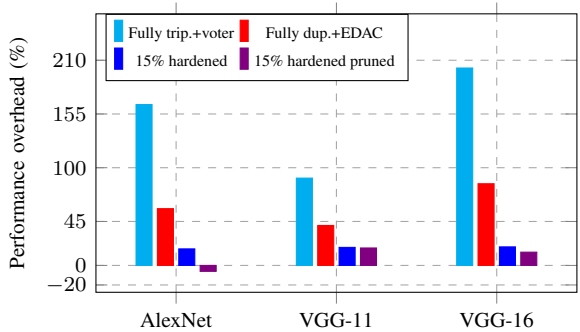
\begin{figure}[h!]
\centering
\captionsetup{justification=centering}
\begin{tikzpicture}
\begin{axis}[
	x tick label style={
		/pgf/number format/1000 sep=},
        font=\footnotesize,
	ylabel=Performance overhead (\%),
	enlargelimits=0.25,
	legend style={at={(0.5,-0.15)},
		anchor=north,
        legend columns=2},
	ybar,
	bar width=8pt,
    ymin=20=,ymax=210,
    xtick={1930, 1940, 1950}, 
    ytick={-20,0,45,100,155,210},
    xticklabels = {AlexNet, VGG-11, VGG-16},
    width=0.9\columnwidth,
    height=0.6\columnwidth,
    legend style={draw=black, at={(0.335,1)}, text opacity = 1,row sep=0pt, font=\fontsize{6}{4}\selectfont},
    grid=major,
    grid style={dashed,gray!70},
        	bar width=6pt
]

\addplot [cyan, fill]
	coordinates {(1930,164.63) (1940,89.37) (1950,202.05)  };
 
\addplot [red, fill]
	coordinates {(1930,58.1) (1940,40.9) (1950,83.6)  };
\addplot [blue, fill]
	coordinates {(1930,16.92) (1940,18.46) (1950,18.97)  };
\addplot [violet, fill]
	coordinates {(1930,-6.06) (1940,17.86)  (1950,13.58)  };

\legend{Fully trip.+voter, Fully dup.+EDAC, 15\% hardened, 15\% hardened pruned}
\end{axis}
\end{tikzpicture}
\caption{Performance overhead comparison for hardened CNNs.}
\label{fig:overhead-cmp}
\end{figure}

Fig. \ref{fig:overhead-cmp} compares the performance overhead in terms of the execution time of different hardened CNNs on NVIDIA 3090 GPU. As observed, the overhead of \textit{triplication + voter} is significantly higher than the other methods. On the other hand, hardened pruned CNNs have the best performance among the hardened CNNs. The resilience of the hardened CNNs is presented in Fig. \ref{fig:dmr-tmr-comparison}-a, Fig. \ref{fig:edac-acc-perf-comparison}, and Fig. \ref{fig:pruned-resilience}.

Throughout the results, the performance of 15\% hardened pruned Alexnet, VGG-11, and VGG-16 is improved by 24\%, 1\%, and 4.7\%, respectively, compared to the 15\% hardened ones without pruning. It is noteworthy that the hardened pruned AlexNet has 6.06\% less execution time than its unprotected baseline.  The selective hardened pruned AlexNet, VGG-11, and VGG-16 require 81.40\%, 2.67\%, and 3.98\% less memory, respectively,  than their unprotected baseline to store their parameters.

\section{Conclusions} \label{sec:conclusion}

This paper presents a model-level hardening method for CNNs by selective channel duplication and EDAC layers. The proposed method enables CNNs to detect and correct faults inherently, at inference time. The hardened CNNs perform reliably at orders of magnitude higher error rates than unprotected CNNs with merely a 15\% hardening ratio, yet incurring 12\% performance overhead. To further minimize the incurred overhead by the hardening method, for the first time, a vulnerability-based pruning that improves resilience is presented. As a result, the hardened pruned CNNs achieve up to 24\% higher performance than the un-pruned hardened CNNs. 

\bibliographystyle{IEEEtran}
\bibliography{refs.bib}

\end{document}